\documentclass[nonacm, sigconf,table]{acmart}
\renewcommand\footnotetextcopyrightpermission[1]{} 
\setcopyright{acmcopyright}
\copyrightyear{2023}
\acmYear{2022}
\acmDOI{XXXXXXX.XXXXXXX}

\usepackage{booktabs}       
\usepackage{scalefnt}
\usepackage{nicefrac}       
\usepackage{microtype}      
\usepackage{optidef}
\usepackage{amsmath,amsfonts}
\usepackage{bbm}
\usepackage{mathtools}
\usepackage{stmaryrd}
\usepackage{subcaption}
\usepackage{graphicx} 
\usepackage{bbm}
\usepackage{paralist}
\usepackage{enumitem}
\setitemize{noitemsep,topsep=0pt,parsep=0pt,partopsep=0pt}

\newtheorem{proposition}{Proposition}

\usepackage[linesnumbered,ruled]{algorithm2e}

\SetKwInput{KwInput}{Input}                
\SetKwInput{KwOutput}{Output}  
\SetKwRepeat{Repeat}{repeat}{until}
\SetKwRepeat{RepeatFor}{repeat}{for}
\SetKwRepeat{Repeat}{repeat}{until}

\newcommand\blfootnote[1]{%
  \begingroup
  \renewcommand\thefootnote{}\footnote{#1}%
  \addtocounter{footnote}{-1}%
  \endgroup
}
\title{FAST: An Optimization Framework for Fast Additive Segmentation in Transparent ML}

\author{Brian Liu}
\affiliation{%
  \institution{Massachusetts Institute of Technology }
   \city{Cambridge}
   \state{Massachusetts}
  \country{USA}}
\email{briliu@mit.edu}

\author{Rahul Mazumder}
\affiliation{%
  \institution{Massachusetts Institute of Technology }
   \city{Cambridge}
   \state{Massachusetts}
  \country{USA}}
\email{rahulmaz@mit.edu}
\renewcommand{\shortauthors}{Brian Liu \& Rahul Mazumder}

\begin{document}

\begin{abstract}

We present FAST, an optimization framework for fast additive segmentation. FAST segments piecewise constant shape functions for each feature in a dataset to produce transparent additive models. The framework leverages a novel optimization procedure to fit these models $\sim$2 orders of magnitude faster than existing state-of-the-art methods, such as explainable boosting machines \citep{nori2019interpretml}. We also develop new feature selection algorithms in the FAST framework to fit parsimonious models that perform well. Through experiments and case studies, we show that FAST improves the computational efficiency and interpretability of additive models.

\end{abstract}

\maketitle

\section{Introduction}

Additive models are popular in machine learning for balancing a high degree of explainability with good predictive performance \citep{ caruana2015intelligible,chang2021interpretable}. These models, when fit on a dataset with $p$ features, take the form $\sum_{j=1}^p s_j(x_j)$. Each additive component $s_j$ is the shape function of feature $x_j$, and since the contribution of each feature can be readily observed from its shape function, additive models are said to be inherently transparent. One such additive model, explainable boosting machines (EBMs), combines this inherent transparency with the powerful predictive performance of tree ensembles \citep{nori2019interpretml}. EBMs use single-feature decision trees, fit via a cyclic boosting heuristic, to construct shape functions. As such, the shape functions built are piecewise constant, a departure from classical and popular smooth components such as those based on polynomials or splines \citep{hastie1995generalized}. Using piecewise constant shape functions, EBMs can capture discontinuities in the underlying data, patterns that are unobserved by smooth additive models and which often have real-world significance \citep{caruana2015intelligible,lengerich2022death}. 
EBMs have also been shown to match the predictive performance of black box methods in various applications while preserving model transparency \citep{chang2021interpretable}. Due to these advantages, EBMs are rapidly becoming ubiquitous in high-stakes applications of ML, such as criminal justice \citep{chang2021interpretable} and healthcare \citep{caruana2015intelligible}, where model explainability is critical.

Inspired by the success of EBMs, and stemming from a reinterpretation of the method, we propose an alternative, FAST. FAST is a formal optimization-based procedure to fit piecewise constant additive models (PCAMs). Both methods construct piecewise constant shape functions, but FAST does so by minimizing a regularized optimization objective while EBMs use a cyclic boosting heuristic.

Moreover, the main goal of FAST is to address the limitations of EBMs that result from this cyclic boosting heuristic. Starting from the null model, EBMs are fit by cycling round-robin over the features and building single-feature decision trees on the boosted residuals, which are dampened by a learning rate. To ensure that the ordering of the features is irrelevant, this learning rate must be kept small. As a result, many cyclic boosting iterations and trees are required to fit an EBM that performs well. This increases the complexity and computational cost of the algorithm and consequently, EBMs struggle to scale for larger datasets.
As a motivating example, consider the UK Black Smoke dataset (9 million rows and 14 columns) used by \cite{wood2017generalized} to test the computational feasibility of splines. It takes the InterpretML package \citep{nori2019interpretml} nearly \textbf{4 hours} to fit an EBM using the default hyperparameters, which are optimized for computation time. FAST, on the other hand, leverages a specialized greedy optimization algorithm to fit a PCAM that performs the same in terms of accuracy in under \textbf{1 minute}. The cyclic heuristic used to fit EBMs also produces feature-dense models by design. This may harm interpretability since an EBM fit on a high dimensional dataset ($p > 50$ features) will contain too many shape functions for a practitioner to explain. FAST introduces two novel feature selection algorithms to remedy this, and these new methods outperform existing feature-sparse PCAM algorithms by up to a \textbf{30\%} reduction in test error. We summarize the contributions of our paper below.

\vspace{1mm}
\textbf{\noindent Main Contributions}

\begin{itemize}[leftmargin=*]
    \item We introduce FAST, an efficient optimization framework to fit PCAMs that supports feature sparsity.

    \item FAST uses a novel procedure to improve computational efficiency. To solve optimization problems in FAST, we apply a computationally cheap greedy block selection rule to an implicit reformulation of our original problem in order to guide a block coordinate descent algorithm. This procedure can fit PCAMs 2 orders of magnitude faster than existing SOTA methods.
    
    
    
    \item We introduce 2 new feature selection algorithms to build sparse PCAMs, an iterative algorithm that relies on our greedy block selection rule and a group $\ell_0$-regularized optimization algorithm.
    
    \item We investigate how correlated features impact feature selection and shape functions in PCAMs and discuss implications for model trustworthiness.
    
\end{itemize}



\vspace{1mm}

We first discuss the advantages of PCAMs over smooth additive models and overview existing algorithms to build PCAMs. Following these preliminaries, we introduce the  FAST optimization framework (\S\ref{framework.section}) and present its novelties: the greedy optimization procedure used to accelerate computation (\S\ref{greedyblockselection.section}) and the feature selection algorithms used to support feature sparsity (\S\ref{fs.section}).

\blfootnote{Supplementary Material: \href{https://github.com/brianliu12437/FAST_segmentation}{github.com/brianliu12437/FAST\_segmentation}}

\subsection{Why PCAMs?} 
\begin{figure}[h!]
  \begin{minipage}[c]{0.28\textwidth}
    \includegraphics[width=\textwidth]{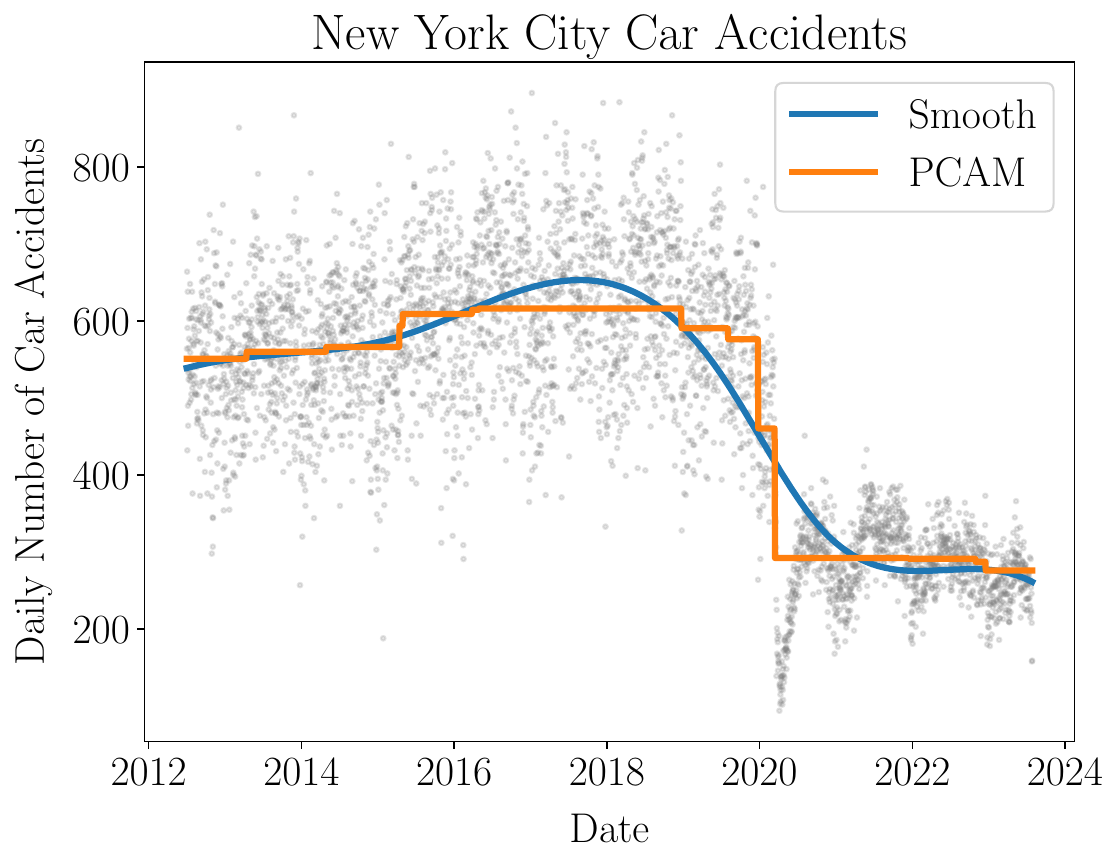}
  \end{minipage}\hfill
  \begin{minipage}[c]{0.19\textwidth}
    \caption{PCAM shape functions can be used to uncover discontinuities in the underlying data.} \label{NYC_car.fig}
  \end{minipage}
\end{figure}

Compared to smooth additive models such as splines, PCAMs have the advantage that they are able to capture discontinuities in the shape functions. These discontinuities can reveal interesting insights about the underlying data. Consider the example shown in Figure \ref{NYC_car.fig}. The scatterplot shows the daily number of car accidents in New York City over a 12-year period and there is a large jump discontinuity in early 2020 due to the COVID-19 pandemic \citep{NYPD_2023}. This discontinuity is captured by the shape function from a PCAM (in blue) but is interpolated and obscured by the smoothing spline (in orange).

The discontinuities observed in PCAM shape functions have been used to uncover hidden patterns in mortality risk models for ICU patients \citep{lengerich2022death} and patients with pneumonia \citep{caruana2015intelligible}. These patterns would have been difficult to detect with smooth additive models or black-box methods. PCAMs also have the advantage that fitted piecewise constant shape functions can be represented by a set of breakpoints. As a result, fitted PCAMs are straightforward to productionize and can be hard-coded into any language with conditional statements (e.g. SQL). Finally, PCAM predictions only require lookups and addition so PCAMs are extremely fast at inference \citep{nori2019interpretml}.

\subsection{Existing PCAM Algorithms:} As mentioned earlier, EBMs use single-feature decision trees, fit via a cyclic boosting heuristic, to build PCAMs \citep{lou2012intelligible}. EBMs are interpretable and perform well, but  are slow to train and feature-dense by design \citep{nori2019interpretml}. Besides EBMs, various methods have been used to construct PCAMs. Additive isotonic models use isotonic regression with backfitting to build PCAMs with monotonic shape functions \citep{bacchetti1989additive}. Spline-based frameworks can also fit PCAMs using zero-degree splines \citep{stone1985additive}. More recently, the fused LASSO has been used to fit PCAMs via ADMM \citep{chu2013distributed} or cyclic block coordinate descent \citep{petersen2016fused}. The latter approach is better known as the fused LASSO additive model (FLAM) and is considered a SOTA algorithm for building PCAMs. As such, we primarily compare FAST against FLAMs and EBMs for fitting feature-dense PCAMs.



\section{FAST Optimization Framework} \label{framework.section}


We introduce FAST and outline the optimization algorithm used to solve problems in our framework. More importantly, we motivate why our greedy optimization procedure (\S\ref{greedyblockselection.section}) improves efficiency.


Given data matrix $X \in \mathbb{R}^{n \times p}$ and target vector $y \in \mathbb{R}^n$, our goal is to fit additive model $\sum_{j=1}^p s_j(x_j)$, where each shape function $s_j$ is piecewise constant. To accomplish this, we introduce a decision variable for each entry in $X$. These decision variables are grouped into decision vectors $\beta_j \in \mathbb{R}^n$ for $j \in [p]$, where each decision vector $\beta_j$ represents the block of decision variables that correspond to feature $x_j$. The decision variables in $\beta_j$ are ordered with respect to the sorted values of $x_j$ and the sum of decision vectors gives the prediction of our model. We fit this prediction to $y$ and recover shape functions $s_j$ from the fitted decision vectors $\beta_j^{*}$. 

\subsection{Optimization Problem}
Let $\beta$ denote the set of  decision vectors $\{ \beta_1 \ldots \beta_p \}$. FAST minimizes the objective $L(y,\beta) + S(\beta)$ to fit PCAMs, where $L$ is a loss function that captures data fidelity and $S$ is a segmentation penalty that encourages piecewise constant segmentation in the fitted decision vectors.  The optimization problem can be written as:\begin{mini}|s|
{\beta_1, \ldots, \beta_p} {\frac{1}{2} \Vert y -  \sum_{j = 1}^p Q_j^\intercal \beta_j\Vert_2^2 + \lambda_f \sum_{j=1}^p \Vert  D \beta_j \Vert_1.}{\label{baseoptimization_problem}}{}
\end{mini}
The first term in the objective is quadratic loss, where $Q_j \in \{0,1\}^{n \times n}$ is the square sorting matrix for feature $x_j$. In other words, $Q_j x_j$ returns the elements of $x_j$ sorted in ascending order and $Q_j^\intercal (Q_j x_j) = x_j$. Since each decision vector $\beta_j$ is ordered with respect to the sorted values of $x_j$, $\sum_{j=1}^p Q_j^\intercal \beta_j$ gives the prediction of our model. The second term in the objective is the fused LASSO segmentation penalty, where $\lambda_f$ is the parameter that controls the number of piecewise constant segments in the shape functions. Higher values of $\lambda_f$ result in less flexible shape functions with fewer segments.
Matrix $D \in \{-1,0, 1\}^{(n-1) \times n}$ is the differencing matrix, where $D \beta_j$ returns a vector of the successive differences of $\beta_j$. 

Problem \ref{baseoptimization_problem} fits feature-dense PCAMs. An optional group sparsity constraint can applied over the blocks $\beta_j$ to select features and we discuss this further in \S\ref{fs.section}.

\subsection{Optimization Algorithm}

Problem \ref{baseoptimization_problem} is convex and separable over blocks $\beta_j$; we develop a block coordinate descent (BCD) algorithm to solve this problem to optimality. Our algorithm has two components: block selections and block updates, and starting with all blocks $\beta_j = 0$ we alternate between the two until convergence.

\textbf{\noindent Block Updates:}
It is critical to note that block updates in FAST are expensive. For a selected block $k$, let $\delta = \{1 \ldots p\} \setminus k$ and define residual vector $r = y - \sum_{j \in \delta} Q_j^\intercal \beta_j$. Each block update solves:
\begin{mini}|s|
 {\beta_k} {\frac{1}{2} \Vert r -   Q_k^\intercal \beta_k\Vert_2^2 + \lambda_f  \Vert  D \beta_k \Vert_1,}
{\label{blockoptimization_obj_problem}}{} 
\end{mini}
which is equivalent to a fused LASSO signal approximation (FLSA) problem on $Q_k r$. These FLSA problems are solved using dynamic programming \cite{johnson2013dynamic} which is computationally expensive.

\textbf{\noindent Block Selections:} Since block updates are expensive, improving the efficiency of our BCD algorithm relies on reducing the number of block updates that we conduct. To do so, we try to select the block that makes the most progress towards the optimal solution in each BCD iteration. Other selection rules, such as cyclic or randomized selection \cite{nutini2022let}, bottleneck BCD with unnecessary updates.

We also must select blocks \emph{cheaply} since block selection would be ineffective if the cost of selecting the best block to update is similar to the cost of updating all blocks. One novelty in FAST is that we develop a greedy optimization procedure to select blocks extremely efficiently. We present this procedure below.

\section{Greedy Optimization Procedure} \label{computation_speed.section}
\label{greedyblockselection.section}





Our greedy optimization procedure hinges on the fact that we can transform Problem \ref{baseoptimization_problem} into an equivalent LASSO problem with $n$ rows and $(n-1)p$ variables. While many LASSO algorithms exist \cite{friedman2007pathwise}, it is infeasible to solve this problem directly since there are too many variables when $n$ is large and the variables are heavily correlated by design \citep{qian2016stepwise}. Rather, we use this LASSO reformulated problem to guide block selection when we apply BCD to Problem \ref{baseoptimization_problem}.

Importantly, we exploit the structure of the design matrix in our LASSO reformulation to derive an extremely efficient block selection rule. In fact, our block selection rule only requires an implicit LASSO reformulation of the original problem (Problem \ref{baseoptimization_problem}), where the design matrix is not explicitly constructed. This is crucial since constructing the design matrix requires a space complexity of $O(n^2p)$, which is infeasible for large data. For example, the design matrix for the UK Black Smoke problem (9 million rows and 14 columns) mentioned in the introduction would take over $10^9$ TB of memory if explicitly constructed.


\subsection{Implicit LASSO Reformulation}  We define a new set of decision vectors $\theta_j \in \mathbb{R}^{n-1}$ for $j \in [p]$, where each vector $\theta_j$ contains the successive differences of vector $\beta_j$. Let $A \in \{0,1\}^{n \times (n-1)}$ be a padded lower triangular matrix with zeros in the first row. We first reformulate Problem \ref{baseoptimization_problem} as:
\begin{mini}|s|
{\theta_1, \ldots, \theta_p} {\frac{1}{2} \Vert y -  \sum_{j = 1}^p Q_j^\intercal A \theta_j\Vert_2^2 + \lambda_f \sum_{j=1}^p \Vert   \theta_j \Vert_1. }{\label{reform1_problem}}{}
\end{mini}
Let $\theta \in \mathbb{R}^{(n-1)p}$ represent the decision vectors $\{\theta_1 \ldots \theta_p\}$ vertically stacked. Let $A' \in \{0,1\}^{np \times (n-1)p}$ be the matrix formed by stacking $A$ submatrices $p$ times along the main diagonal. Let $Q^\intercal \in \{0,1\}^{np \times np}$ be the matrix formed by stacking $\{Q_1^\intercal \ldots Q_p^\intercal\}$ along the main diagonal. Finally, let $M \in \{0,1\}^{n \times np}$ be the matrix formed by stacking $p$ identity matrices of dimension $n \times n$ horizontally. We show a visualization of these matrices in the appendix (suppl. A). Problem \ref{reform1_problem} and Problem \ref{baseoptimization_problem} are equivalent to:
\begin{mini}|s|
{\theta} {\frac{1}{2} \Vert y -  MQ^\intercal A' \theta\Vert_2^2 + \lambda_f  \Vert \theta \Vert_1, }{\label{reform2_problem}}{}
\end{mini}
which is a block-separable LASSO problem with design matrix $MQ^\intercal A' \in \mathbb{R}^{n \times (n-1)p}$. We show in the next section that we do not need to construct this matrix for our greedy selection rule.



\subsection{Block Selection Rule (BGS rule)} \label{bgs_rule_sub.section}

Since Problem \ref{reform2_problem} is an equivalent LASSO reformulation of Problem \ref{baseoptimization_problem}, we use this reformulation to select which blocks to update when performing BCD on Problem \ref{baseoptimization_problem}. For each BCD iteration, we apply a block Gauss Southwell (BGS) greedy selection rule to Problem \ref{reform2_problem} to select the next block to update. BGS selection has been shown in theory and in practice to make more progress per iteration than cyclic or random selection \cite{nutini2015coordinate,dhillon2011nearest}, however, on many problems, BGS selection is prohibitively expensive \cite{nutini2022let}. One critical aspect of our procedure is that we exploit problem structure to develop a BGS steepest direction (BGS-s) rule that is cheap to compute.

Let $f(\theta) = \frac{1}{2} \Vert y -  MQ^\intercal A' \theta\Vert_2^2$. For BGS-s selection, we first  compute vector $d \in \mathbb{R}^{np}$ which stores the magnitude of the most negative directional derivative for each coordinate. This vector is defined coordinate-wise by
\begin{equation}\label{steepest_direction_vector}
    d_i = \begin{cases} |S_{\lambda_f}(\nabla_i f(\theta)| &\text{if} \  \theta_i = 0 \\ |\nabla_i f(\theta) + \text{sign}(\theta_i)\lambda_f| & \text{if} \ \theta_i \neq 0,
\end{cases}
\end{equation}
where $S_{\lambda_f}$ is the soft-thresholding operator. Let  $d_k \in \mathbb{R}^{n}$ represent the elements in vector $d$ associated with block $k$. We select the best block $k^*$ to update via:
 \begin{argmaxi}|s|
{k \in [p] }{\Vert d_{k} \Vert_2^2.}{\label{steepest_block_selection}}{k^* = }
 \end{argmaxi}

Equations \ref{steepest_direction_vector} and \ref{steepest_block_selection} form our BGS selection rule, which is computationally bottlenecked by the cost of computing the full gradient $\nabla f(\theta)$. The LASSO design matrix $MQ^\intercal A'$ is also only used to compute this gradient. Below, we show how to efficiently compute gradient $\nabla f(\theta)$ without forming the LASSO design matrix.

\textbf{\noindent Fast Gradient Procedure:} We have that  $\nabla f(\theta) = - {A'}^\intercal Q M^\intercal r'$, where $r' = y - MQ^\intercal A' \theta$. Since our algorithm is zero-initialized, we can store $r'$ and update the residual vector at each BCD iteration to avoid multiplying the design matrix with $\theta$. Matrix $M^\intercal$ consists of $p$ identity matrices stacked vertically which makes the gradient expression block-separable. For a fixed block $k \in [p]$, we have that $\nabla_k f(\theta) = -A^\intercal Q_k r'$, where $Q_k$ is the sorting matrix  for feature $x_j$. The matrix $A^\intercal$ is a padded upper triangular matrix, so computing the gradient for block $k$ simply involves ordering $r'$ with respect to the sorted values of $x_j$ and taking a rolling sum down the ordered vector, which is extremely efficient. Computing the full gradient can be embarrassingly parallelized across blocks. 

With this procedure, our BGS-s selection rule is efficient, parallelizable, and can be computed without constructing the LASSO design matrix. Below, we formalize our greedy block coordinate descent (GBCD) algorithm and analyze its convergence properties.


\subsection{BGS-GBCD Algorithm}

To solve Problem \ref{baseoptimization_problem}, we use the following GBCD algorithm. Start with $\beta_j = 0$ for all blocks $j \in [p]$ and repeat until convergence: apply our BGS selection rule to Problem \ref{reform2_problem} (LASSO reformulation) to select a block to update and solve Problem \ref{blockoptimization_obj_problem} (original block update problem) with dynamic programming to update the block. This returns a sequence of solutions $\beta^{t}$ that correspond to a sequence of decreasing objective values.

\subsubsection{Convergence Analysis} The sequence of solutions $\beta^t$ returned by BGS-GBCD converges to the minimizer for Problems \ref{baseoptimization_problem} and \ref{reform2_problem}. More generally, we show that BGS-GBCD converges to optimality when applied to block-separable LASSO problems. We prove the next proposition in the appendix (suppl. B.1).
\begin{proposition} Given composite problem
\begin{mini*}|s|
{\theta} {F(\theta) = f(\theta) + \lambda \Vert \theta \Vert_1,}{\label{general_problem}}{}
\end{mini*}
where $f$ is convex and coordinate-wise $\mathcal{L}$-smooth and $\theta$ is both block and coordinate separable, every limit point of BGS-GBCD coincides with a minimizer for $F(\theta)$. Any sequence of solutions $\theta^t$ generated by BGS-GBCD converges to a limit/minimum point.
\end{proposition}
We prove for the first time that greedy block coordinate descent using block Gauss-Southwell-s selection converges to the minimum point when applied to $\ell_1$-composite problems. We also show that under certain conditions, BGS-GBCD updates make provably good progress towards the minimum. Proposition \ref{proposition2} states a property that we exploit in \S\ref{fs.section} when developing feature selection algorithms.
\begin{proposition}\label{proposition2}
If block $\theta_k^t = \textbf{0}$ is selected via the BGS rule, the progress after one GBCD  update is bound by:
\begin{mini*}|s|
{\gamma \in \mathbb{R}^n}{\nabla f(\theta^t)^\intercal \gamma \ +  \breakObjective{\frac{\mathcal{L}}{2} \text{snorm}(\gamma) +\lambda \Vert \theta^t + \gamma \Vert_1 - \lambda \Vert \theta^t \Vert_1,}}{\label{steepest_direction_vector_thm}}{F(\theta^{t+1}) - F(\theta^t) \leq }
 \end{mini*}
where snorm($\gamma$) sums the $\ell_2$-norm of each block in $\gamma$.
\end{proposition}

Each block in our optimization problem corresponds to a feature; $\beta_j$ gives the contribution of feature $x_j$ to the additive model.
Proposition \ref{proposition2} states that when the BGS rule is used to select a feature (block) to enter the support, the corresponding block update makes substantial progress towards the minimum. The proof for this proposition is also in the appendix (suppl. B.2).



\subsection{Discussion} In most LASSO problems, greedy selection offers little advantage over cyclic selection since the computational cost of selecting the block with the steepest directional derivatives is similar to the cost of updating all of the blocks \citep{wu2008coordinate}. Greedy BCD is effective in FAST, however, since block selections, which involve embarrassingly parallel summations are much cheaper than block updates, which require expensive dynamic programming calls.

We observe that BGS-GBCD greatly reduces the number of dynamic programming block updates (Problem \ref{blockoptimization_obj_problem}) required to solve Problem \ref{baseoptimization_problem}, compared to cyclic block selection. 
\begin{figure}[h!]
  \begin{minipage}[c]{0.28\textwidth}
    \includegraphics[width=\textwidth]{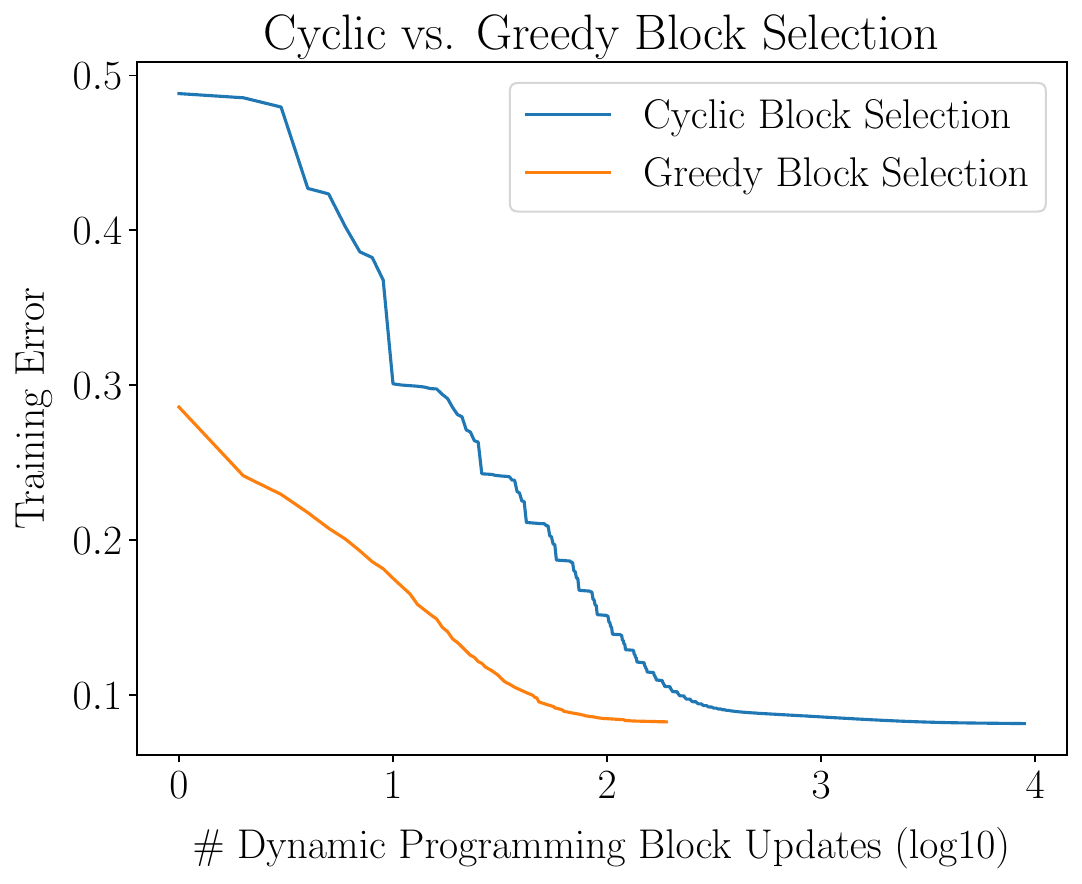}
  \end{minipage}\hfill
  \begin{minipage}[c]{0.175\textwidth}
    \caption{Greedy selection reduces the number of dynamic programming block updates by 2 orders of magnitude.} \label{greedy_v_cyclic.fig}
  \end{minipage}
\end{figure}
For example, in Figure \ref{greedy_v_cyclic.fig}, we use greedy and cyclic BCD to fit FAST on the Elevators dataset \citep{vanschoren2014openml}. The horizontal axis shows the number of dynamic programming block updates (log10) and the vertical axis shows training loss. We observe that BGS-GBCD requires nearly 100$\times$ fewer updates to converge. This corresponds to substantial computational speedups, which we show in our experiments in  \S\ref{computation_time_experiment.section}.

\subsection{Binning} FAST can also incorporate binning, a popular heuristic used by EBMs \citep{nori2019interpretml} and LightGBMs \citep{ke2017lightgbm}, to reduce computation time for a nominal cost in model expressiveness. FAST performs binning using a novel equivalent optimization formulation while existing methods, such as EBMs, pre-process the data.
Given a set of bins for each feature $x_j$, we add the constraints for all entries $(i_1, i_2) \in [n]$ that if entries $(x_j)_{i_1}$ and $(x_j)_{i_2}$ fall in the same bin, then $(\beta_j)_{i_1} = (\beta_j)_{i_2}$. We show in the appendix (suppl. C) that we can reformulate these constraints into a weighted smooth loss function in the objective and efficiently solve this unconstrained problem with BGS-GBCD. Binning directly reduces the number of decision variables in FAST by a factor of \# bins over \# rows and combining BGS-GBCD with binning further reduces computation time.

\section{Feature-Sparse FAST} \label{fs.section}

Our FAST framework is quite flexible; here we discuss an extension of the framework to explicitly account for variable selection. We add this group sparsity constraint to Problem \ref{baseoptimization_problem}: $\sum_{j=1}^p \mathbbm{1}(\beta_j \neq \textbf{0}) \leq K$, where $K$ is the maximum number of features to select. Problem \ref{baseoptimization_problem} with this constraint is NP-hard and difficult to solve to optimality due to the large number of variables; we have a variable for each entry of $X$. As such, we develop two approximate algorithms to find good solutions. These algorithms have different strengths in terms of solution quality and runtime, but both algorithms rely on the BGS rule presented in \S\ref{bgs_rule_sub.section} and the fact that BGS selection makes provably good progress when selecting features to enter the support (Prop. \ref{proposition2}).


\subsection{Approximate Greedy Iterative Selection}\label{agis_section}

For Approximate Greedy Iterative Selection (AGIS), we partition the blocks into the  support $S = \{ j \in [p] \mid \beta_j \neq \bf{0} \}$ and complement $S^c$ and start with all blocks equal to $\bf{0}$. We use the BGS rule to select the best block $k \in S^c$ to update and we perform a block update by solving Problem \ref{blockoptimization_obj_problem} to add $k$ into $S$. If $|S| > 1$, we iterative through the blocks in $S$ and conduct block updates until convergences. We repeat this procedure, interlacing BGS selection with sweeps on the support $S$ until the condition $|S| = K$ is reached. AGIS returns a sequence of PCAMs with every feature sparsity level from $1 \dots K$. To improve solution quality across all sparsity levels we apply this local search heuristic.

\subsubsection{BGS Local Search:}\label{localsearch.section} After each sweep of $S$ converges, use the BGS rule to select the best block to update in $S^c$ and denote that block $\beta_j^*$. This is the block that we will swap into the support. To find the best block to swap out of the support, iterate over $\beta_j \in S$. For each block, set $\beta_j = \textbf{0}$ and conduct a block update on $\beta_j^*$, and select the block in $S$ that when swapped improves the objective the most. After this swap, conduct another sweep over $S$ until convergence to obtain the final solution. We present our full AGIS algorithm, with local search, in Algorithm \ref{agisalgo}.

\begin{algorithm}[h]
\scriptsize
\caption{AGIS}
\label{agisalgo}
\DontPrintSemicolon
  \KwInput{$K$, $\lambda_f$, $D$, $Q_j  \mkern9mu \forall \ j \in [p]$}
  
  \textbf{Initialize} $\beta_j = \textbf{0} \mkern9mu \forall \ j \in [p]$, $S = \emptyset$ , $S_{all} = \emptyset$
  
  \Repeat{$|S| = K$}{

  Use BGS rule to select $k \in S^c$.

  Update block $k$ (Problem \ref{blockoptimization_obj_problem}).

  $S = S \cup \beta_k, \ S^c = S^c \setminus  \beta_k$
  
  \Repeat{converged}{
Sweep through $S$ and update blocks (Problem \ref{blockoptimization_obj_problem}).
  }
  
  BGS local search.

  $S_{all} = S_{all} \cup S$
  
  }
  \KwOutput{Sequence of models $S_{all}$}

\end{algorithm}

\subsection{Group $\ell_0$-FAST}

In addition to AGIS, we can use a group $\ell_0$-sparsity penalty to select features in FAST. This approach often obtains better solutions at the cost of increased computation time, which we discuss in \S\ref{feature_sparse_experiments.section}. We use this Lagrangian formulation:
\begin{mini}|s|
{\beta} {\frac{1}{2} \Vert y -  \sum_{j = 1}^p Q_j^\intercal \beta_j\Vert_2^2 + \breakObjective{\lambda_f \sum_{j=1}^p \Vert  D \beta_j \Vert_1 + \lambda_s \sum_{j = 1}^p \mathbbm{1}(\beta_j \neq \textbf{0}),\label{groupl0_problem}}}
{}{}
\end{mini}
where $\lambda_s$ is the sparsity hyperparameter.The group sparsity penalty is block-separable over $\beta_j$ so we can apply BCD methods to find good solutions to this problem.
Given fixed block $k$ and residual vector $r$, we can write each block update problem as:
\begin{mini*}|s|
{\beta_k} {\frac{1}{2} \Vert Q_k r - \beta_k\Vert_2^2 + \lambda_f \Vert  D \beta_k \Vert_1 + \lambda_s  \mathbbm{1}(\beta_k \neq \textbf{0}).\label{}}
{}{}
\end{mini*}
This problem can be solved by first setting $\lambda_s = 0$ and solving the FLSA for $\beta_k^*$. We then check the thresholding condition:
\begin{equation*}
   \frac{1}{2}\Vert r \Vert_2^2 - \frac{1}{2}\Vert Q_k r - \beta_k^* \Vert_2^2 - \lambda_f \Vert D \beta_k^* \Vert_1 \leq \lambda_s
\end{equation*}
and set $\beta_k^* = 0$ if the condition is satisfied. We show the derivation for this  in the appendix (suppl. D).

Since the group sparsity penalty is not continuous, it is not clear if BGS-GBCD can be extended here. 
 To find high-quality solutions to Problem \ref{groupl0_problem}, we use cyclic block coordinate descent and apply our BGS local search heuristic (\S\ref{localsearch.section}) when CBCD converges. We interlace CBCD sweeps with local search steps until the objective no longer improves.


\subsection{Discussion}

We show an example of the impact of local search on solution quality and discuss the strengths and weaknesses of both group $\ell_0$-FAST and AGIS.

\begin{figure}[h]
    \centering
    \includegraphics[width = 0.45\textwidth]{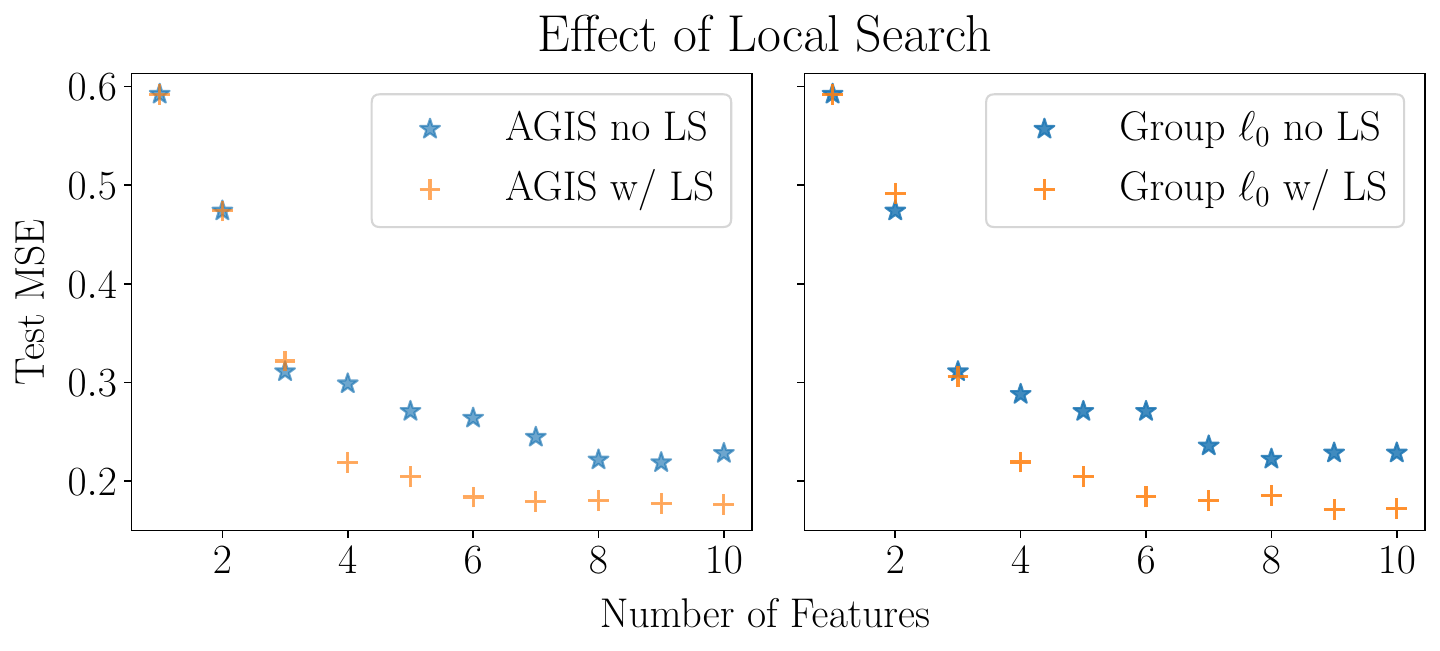}
    \caption{BGS local search improves the solution quality for both of our feature-sparse PCAM algorithms.}
    \label{local_search.fig}
\end{figure}

\subsubsection{Local Search Performance}
We observe empirically that our BGS local search heuristic improves the out-of-sample performance of both feature selection algorithms. For example, in Figure \ref{local_search.fig}, we use group $\ell_0$-FAST and AGIS to build feature-sparse PCAMs on the Elevators dataset (16500 rows and 16 columns) \citep{vanschoren2014openml}. We vary $K$, the sparsity budget in the model from 1 to 10, and compare the test performance of the model measured via MSE. For both methods, the local search heuristic improves performance. 

\subsubsection{Group $\ell_0$-FAST vs. AGIS} \label{agis_vl0.section} In our experiments in \S\ref{feature_sparse_experiments.section}, we observe that group $\ell_0$-FAST generally outperforms AGIS at building sparse PCAMs. AGIS, however, is computationally faster since the algorithm can leverage greedy block selection. Fitting group $\ell_0$-FAST requires CBCD updates due to the non-convexity of the group sparsity penalty. In addition, AGIS is easier to use since the algorithm by design outputs a sequence of PCAMs with with every support size from $1$ to $K$. The sparsity hyperparameter $\lambda_s$ in group $\ell_0$-FAST must be tuned and the algorithm may skip certain support sizes due to non-convexity \citep{hazimeh2020fast}.


\section{Experiments}
We evaluate the computation time of FAST against existing algorithms and assess how well the framework performs at building feature-sparse PCAMs.

\subsection{Computation Time Experiment} \label{computation_time_experiment.section}

We compare the computation time of FAST against existing SOTA algorithms for building feature-dense PCAMs: EBMs and FLAMs.

\subsubsection{Experimental Procedure} On 10 large regression benchmark datasets from OpenML \cite{vanschoren2014openml}, we use FAST, EBM, and FLAM to fit PCAMs. For the competing methods, we use the InterpretML package \citep{nori2019interpretml} to fit EBMs in Python and the FLAM package in R \cite{flampackage}. We use the default hyperparameters for InterpretML EBMs, which are optimized for fast runtime. For FLAM, we match the fusion hyperparameter with the value used in FAST. The test errors of the models fit using the 3 methods, under these configurations, are comparable (as intended). We conduct this experiment on a M2 Macbook Pro with 10 cores and match the number of cores used in the methods that support multiprocessing (FAST and EBMs). Additional details can be found in the appendix (suppl. E).

\begin{table}[h]
\scalebox{0.85}{
\begin{tabular}{|
>{\columncolor[HTML]{EFEFEF}}c |
>{\columncolor[HTML]{FFFFFF}}c |
>{\columncolor[HTML]{FFFFFF}}c |
>{\columncolor[HTML]{FFFFFF}}c |}
\hline
\textbf{Dataset / Method}                                                                   & \cellcolor[HTML]{EFEFEF}\textbf{FAST}                                              & \cellcolor[HTML]{EFEFEF}\textbf{EBM}                          & \cellcolor[HTML]{EFEFEF}\textbf{FLAM}                         \\ \hline
\textbf{\begin{tabular}[c]{@{}c@{}}Black Smoke +\\ (9214951, 41)\end{tabular}}     & {\color[HTML]{000000} \begin{tabular}[c]{@{}c@{}}329.6s (1.2)\\ 0.37\end{tabular}} & \begin{tabular}[c]{@{}c@{}}15h 49m 31s\\ 0.37\end{tabular}    & \_\_                                                          \\ \hline
\textbf{\begin{tabular}[c]{@{}c@{}}Black Smoke\\ (9214951, 14)\end{tabular}}       & {\color[HTML]{000000} \begin{tabular}[c]{@{}c@{}}43s (2.8)\\ 0.38\end{tabular}}    & \begin{tabular}[c]{@{}c@{}}3h 57m 9s\\ 0.38\end{tabular}      & \_\_                                                          \\ \hline
\textbf{\begin{tabular}[c]{@{}c@{}}Physiochemical\\ (5023496, 9)\end{tabular}}     & {\color[HTML]{000000} \begin{tabular}[c]{@{}c@{}}33.4s (0.3)\\ 0.52\end{tabular}}  & \begin{tabular}[c]{@{}c@{}}43m 31s (31.9)\\ 0.53\end{tabular} & \_\_                                                          \\ \hline
\textbf{\begin{tabular}[c]{@{}c@{}}Auto Horsepower\\ (900000, 17)\end{tabular}}    & {\color[HTML]{000000} \begin{tabular}[c]{@{}c@{}}1.63s (0.01)\\ 0.42\end{tabular}} & \begin{tabular}[c]{@{}c@{}}85s (3.9)\\ 0.43\end{tabular}      & \_\_                                                          \\ \hline
\textbf{\begin{tabular}[c]{@{}c@{}}Ailerons BNG\\ (669994, 38)\end{tabular}}       & {\color[HTML]{000000} \begin{tabular}[c]{@{}c@{}}2.57s (0.08)\\ 0.44\end{tabular}} & \begin{tabular}[c]{@{}c@{}}85s (2.0)\\ 0.44\end{tabular}      & \_\_                                                          \\ \hline
\textbf{\begin{tabular}[c]{@{}c@{}}Slice Localization\\ (35845, 351)\end{tabular}} & {\color[HTML]{000000} \begin{tabular}[c]{@{}c@{}}6.7s (0.05)\\ 0.18\end{tabular}}  & \begin{tabular}[c]{@{}c@{}}58.7s (2.2)\\ 0.20\end{tabular}    & \begin{tabular}[c]{@{}c@{}}14m 50s (30.4)\\ 0.18\end{tabular} \\ \hline
\textbf{\begin{tabular}[c]{@{}c@{}}Superconduct\\ (21263, 79)\end{tabular}}        & {\color[HTML]{000000} \begin{tabular}[c]{@{}c@{}}0.45s (0.01)\\ 0.20\end{tabular}} & \begin{tabular}[c]{@{}c@{}}7.6s (0.05)\\ 0.21\end{tabular}    & \begin{tabular}[c]{@{}c@{}}13.0s (0.03)\\ 0.20\end{tabular}   \\ \hline
\textbf{\begin{tabular}[c]{@{}c@{}}Scm1d\\ (8828, 280)\end{tabular}}               & \begin{tabular}[c]{@{}c@{}}0.7s (0.01)\\ 0.10\end{tabular}                         & \begin{tabular}[c]{@{}c@{}}7.9s (0.47)\\ 0.12\end{tabular}    & \begin{tabular}[c]{@{}c@{}}190s (5.1)\\ 0.10\end{tabular}     \\ \hline
\textbf{\begin{tabular}[c]{@{}c@{}}Rf2\\ (8212, 448)\end{tabular}}                 & \begin{tabular}[c]{@{}c@{}}1.34s (.01)\\ 0.02\end{tabular}                         & \begin{tabular}[c]{@{}c@{}}78.5s (4.3)\\ 0.02\end{tabular}    & \begin{tabular}[c]{@{}c@{}}180s (3.0)\\ 0.03\end{tabular}     \\ \hline
\textbf{\begin{tabular}[c]{@{}c@{}}Isolet\\ (7017, 613)\end{tabular}}              & \begin{tabular}[c]{@{}c@{}}2.50s (0.02)\\ 0.32\end{tabular}                        & \begin{tabular}[c]{@{}c@{}}8.67s (0.3)\\ 0.32\end{tabular}    & \begin{tabular}[c]{@{}c@{}}360s (8.0)\\ 0.34\end{tabular}     \\ \hline
\end{tabular}}
\caption{Timing experiment results. FAST achieves 2 orders of magnitude speedups for large problems.}
\label{timing.table}
\end{table}

\subsubsection{Results} Table \ref{timing.table} shows the results of our experiment. The leftmost column shows dataset names and dimensions: $(n,p)$. In each cell in the other columns, the top entry shows the computation time of the method averaged over runs along with the standard deviation. The bottom entry shows the test MSE of the model. 

The top 5 rows of this table show timing results on large $n$ datasets with more than $500000$ rows. On these datasets, we are unable to apply FLAM due to problem scale so we compare FAST against EBMs. We observe that FAST fits PCAMs up to 2 orders of magnitude faster than EBMs. For example on an augmented version of the UK Black Smoke dataset, with 9 million rows and 41 columns, it takes over 15 hours to fit an EBM. FAST on the other hand can fit a PCAM that performs the same in around 5 minutes.


The bottom 5 rows of Table \ref{timing.table} show results on large $p$ datasets with more than 50 columns, but less than 50,000 rows. We observe here that FAST fits PCAMs 2 orders of magnitude faster than FLAM and around 1 order of magnitude faster than EBMs. For example on the Slice Localization dataset, with over 300 columns, it takes nearly 15 minutes to fit a FLAM. FAST can fit a PCAM that performs the same in under 10 seconds. 

In all, we find that FAST substantially outperforms EBMs and FLAMs in terms of computation time across various large problems.

\subsubsection{Low Optimization Tolerance Models} 

\begin{figure}[h]
    \centering
    \includegraphics[width = 0.46\textwidth]{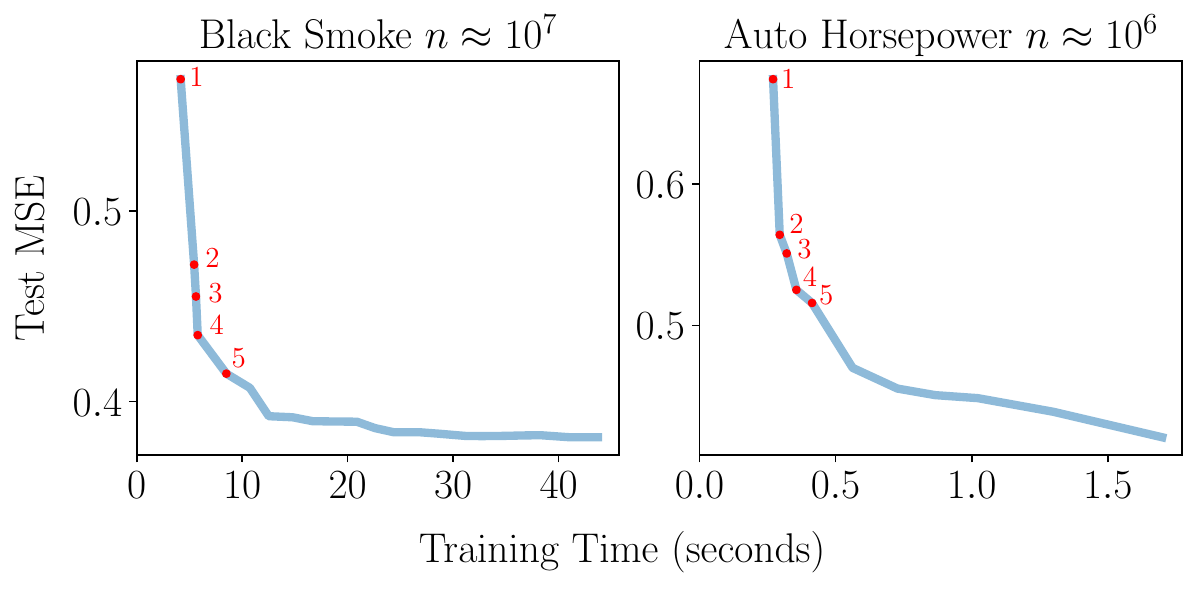}
    \caption{FAST can fit low optimization tolerance models that perform well extremely quickly.}
    \label{timing_budget.pdf}
\end{figure}

As an aside, we note that we can leverage our greedy BCD algorithm to fit FAST with low optimization tolerances, in order to quickly produce a PCAM that still performs well out-of-sample.  In Figure \ref{timing_budget.pdf}, we show the test error of FAST (vertical axes) plotted against the training time in seconds (horizontal axes) for the UK Black Smoke and Auto Horsepower \cite{vanschoren2014openml} datasets. We vary the training time of FAST by early-stopping the optimization algorithm after a fixed number of iterations, the first 5 GBCD iterations are plotted in red. In both examples, the first 5 iterations greatly reduce the \emph{test} error of the model. For the UK Black Smoke dataset, FAST can fit a low optimization tolerance model that performs well in less than 10 seconds.



\subsection{Feature Selection Experiment} \label{feature_sparse_experiments.section}

Here we evaluate how well FAST performs at building feature-sparse PCAMs.

\subsubsection{Experimental Setup} We repeat this procedure on 20 regression datasets from OpenML and use a 10-fold CV on each dataset. The full list of datasets can be found in the appendix (suppl. E). On the training folds, we use group $\ell_0$-FAST and AGIS to fit feature-sparse PCAMs by varying the sparsity budget $K \in \{2, 4, 6, 8, 10\}$. We evaluate the MSE of each sparse model on the test fold. 

We compare the performance of these models against the following SOTA algorithms to construct feature-sparse PCAMs.

\vspace{1mm}
\begin{itemize}[leftmargin=*]

\item \textbf{FLAM-GL (2016):} In FLAM group LASSO \cite{petersen2016fused}, we fit a FLAM with a group LASSO penalty over the features. We tune the sparsity hyperparameter such that at most $K$ features are selected.

\item \textbf{EBM-RS (2019):} In EBM rank and select \cite{nori2019interpretml}, we first fit an EBM on the training data and rank the features by importance scores; the contribution of each feature averaged over the training observations. We select the top $K$ features and refit an EBM. This method is computationally expensive since it fits two PCAMs.

\item \textbf{ControlBurn (2021):} ControlBurn \cite{liu2021controlburn} is a flexible framework for building feature-sparse nonlinear models. The feature selection algorithm in the framework first constructs a specialized tree ensemble that is diverse, where each tree in the ensemble uses a different subset of features. Then, the weighted LASSO is used to select feature-sparse subsets of trees that perform well. We refit the final model, in this case, an EBM, on the $K$ selected features. ControlBurn with an EBM has been used to construct high-performing, feature-sparse PCAMs for heart failure prediction in clinical machine learning \citep{van2023interpretable}.

\item \textbf{FastSparseGAM (2022):} FastSparseGAM \cite{Liu_2023} is a package for sparse regression built on top of the L0Learn framework \cite{hazimeh2020fast, hazimeh2023l0learn}. The package can be adapted to construct \emph{extremely} sparse PCAMs by one-hot encoding the features and selecting a small subset of the resulting components \cite{liu2022fast}.  
\end{itemize}
\vspace{1mm}
We also compare feature-sparse FAST against two traditional algorithms that produce non-piecewise constant additive models, Sparse Additive Models (\textbf{SAM}), which uses the group LASSO to sparsify splines, and the linear \textbf{LASSO}. Additional details on our experimental procedure can be found in the appendix (suppl. E).

\begin{figure*}[h]
    \centering
    \includegraphics[width = 0.97\textwidth]{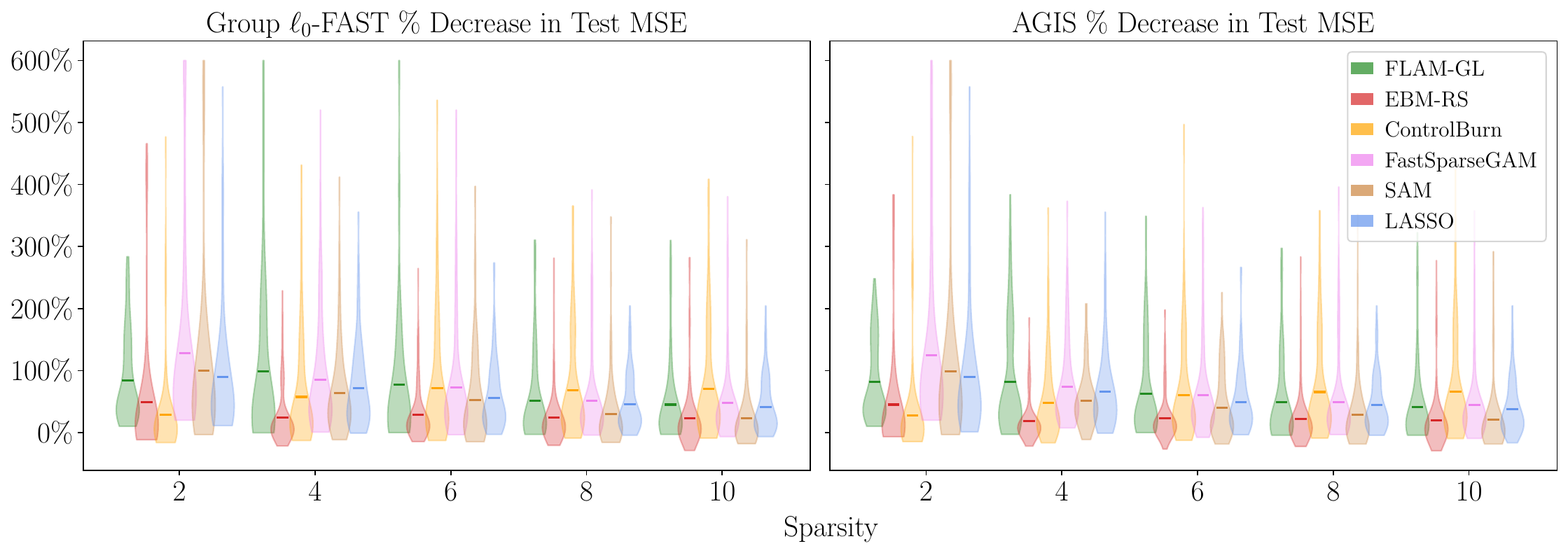}
    \caption{Distribution of results from our feature selection experiment (averages shown in Table \ref{fast_fs.table}). The distributions are mostly entirely positive, which indicates that feature-sparse FAST outperforms our competing algorithms.}
    \label{violin_plot.fig}
\end{figure*}

\subsubsection{Results} For each run of our experiment, we compute the percent decrease in test MSE between feature-sparse FAST and each competing algorithm, given by: \begin{equation*}
 \text{\% decrease MSE} = \frac{\text{MSE Competing Alg.} - \text{MSE FAST}}{\text{MSE FAST}},
\end{equation*} for each sparsity budget. A positive percent decrease in test error indicates that feature-sparse FAST performs better than the competing algorithm for that sparsity budget. 

In Table \ref{fast_fs.table} we report the average percent decrease in test error between group $\ell_0$-FAST and the competing algorithms (\textbf{top} values) and AGIS and the competing algorithms (\textbf{bottom} values) across all sparsity budgets. 
\begin{table}[]
\scalebox{0.85}{
\begin{tabular}{|c|c|c|c|c|c|}
\hline
\rowcolor[HTML]{EFEFEF} 
{\color[HTML]{000000} \textbf{\begin{tabular}[c]{@{}c@{}}Competing Alg. /\\ Sparsity \end{tabular}}} & \textbf{2}                                                & \textbf{4}                                               & \textbf{6}                                              & \textbf{8}                                              & \textbf{10}                                             \\ \hline
\cellcolor[HTML]{A0E8A0}{\color[HTML]{000000} \textbf{FLAM-GL}}                                        & \begin{tabular}[c]{@{}c@{}}84.1\%\\ 81.5\%\end{tabular}   & \begin{tabular}[c]{@{}c@{}}102.0\%\\ 81.6\%\end{tabular} & \begin{tabular}[c]{@{}c@{}}77.9\%\\ 62.3\%\end{tabular} & \begin{tabular}[c]{@{}c@{}}50.8\%\\ 48.6\%\end{tabular} & \begin{tabular}[c]{@{}c@{}}45.1\%\\ 41.4\%\end{tabular} \\ \hline
\cellcolor[HTML]{F1B3B3}{\color[HTML]{000000} \textbf{EBM-RS}}                                         & \begin{tabular}[c]{@{}c@{}}48.9\%\\ 45.1\%\end{tabular}   & \begin{tabular}[c]{@{}c@{}}24.3\%\\ 18.3\%\end{tabular}  & \begin{tabular}[c]{@{}c@{}}29.7\%\\ 22.7\%\end{tabular} & \begin{tabular}[c]{@{}c@{}}23.7\%\\ 22.0\%\end{tabular} & \begin{tabular}[c]{@{}c@{}}22.8\%\\ 19.8\%\end{tabular} \\ \hline
\cellcolor[HTML]{FFE0A6}{\color[HTML]{000000} \textbf{ControlBurn}}                                    & \begin{tabular}[c]{@{}c@{}}28.4\%\\ 27.7\%\end{tabular}   & \begin{tabular}[c]{@{}c@{}}57.5\%\\ 47.5\%\end{tabular}  & \begin{tabular}[c]{@{}c@{}}71.3\%\\ 60.1\%\end{tabular} & \begin{tabular}[c]{@{}c@{}}68.1\%\\ 65.4\%\end{tabular} & \begin{tabular}[c]{@{}c@{}}70.4\%\\ 65.6\%\end{tabular} \\ \hline
\cellcolor[HTML]{F8CDF8}{\color[HTML]{000000} \textbf{FastSparseGAM}}                                  & \begin{tabular}[c]{@{}c@{}}131.5\%\\ 126.5\%\end{tabular} & \begin{tabular}[c]{@{}c@{}}85.6\%\\ 73.4\%\end{tabular}  & \begin{tabular}[c]{@{}c@{}}72.4\%\\ 60.7\%\end{tabular} & \begin{tabular}[c]{@{}c@{}}50.9\%\\ 48.6\%\end{tabular} & \begin{tabular}[c]{@{}c@{}}48.2\%\\ 44.2\%\end{tabular} \\ \hline
\cellcolor[HTML]{EBCEB2}{\color[HTML]{000000} \textbf{SAM}}                                           & \begin{tabular}[c]{@{}c@{}}103.5\%\\ 102.4\%\end{tabular} & \begin{tabular}[c]{@{}c@{}}64.0\%\\ 51.7\%\end{tabular}  & \begin{tabular}[c]{@{}c@{}}52.8\%\\ 40.4\%\end{tabular} & \begin{tabular}[c]{@{}c@{}}30.0\%\\ 28.4\%\end{tabular} & \begin{tabular}[c]{@{}c@{}}23.5\%\\ 20.5\%\end{tabular} \\ \hline
\cellcolor[HTML]{C1D5F8}{\color[HTML]{000000} \textbf{LASSO}}                                          & \begin{tabular}[c]{@{}c@{}}89.6\%\\ 89.5\%\end{tabular}   & \begin{tabular}[c]{@{}c@{}}71.9\%\\ 65.8\%\end{tabular}  & \begin{tabular}[c]{@{}c@{}}56.1\%\\ 49.1\%\end{tabular} & \begin{tabular}[c]{@{}c@{}}46.0\%\\ 44.1\%\end{tabular} & \begin{tabular}[c]{@{}c@{}}41.2\%\\ 38.2\%\end{tabular} \\ \hline
\end{tabular}
}
\caption{Average \% decrease in test error between feature-sparse FAST and our competing algorithms across sparsity budgets (distributions shown in Figure \ref{violin_plot.fig}). Positive values indicate that feature-sparse FAST outperforms the competing algorithm. In each cell, the top value shows group $\ell_0$-FAST and the bottom value shows AGIS. 
\label{fast_fs.table}}
\vspace*{-10mm}
\end{table}
These averages are taken across all datasets and folds in our experiment. In Figure \ref{violin_plot.fig}, we show the full distributions of our results. In each plot, the horizontal axis shows the sparsity budget and the vertical axis shows the percent decrease in test error between feature-sparse FAST and the competing algorithm; the left plot shows group $\ell_0$-FAST and the right plot shows AGIS. The grouped violin plots show the distribution of the results for each sparsity budget and the averages of each distribution are marked by horizontal lines, which correspond to the averages in Table \ref{fast_fs.table}.

From Figure \ref{violin_plot.fig}, we see that group $\ell_0$-FAST and AGIS largely outperform all of our competing algorithms. The distributions of the percent decrease in test error between these two methods and our competing algorithms are nearly entirely positive across all sparsity budgets. We also observe that group $\ell_0$-FAST consistently performs slightly better than AGIS. In Table \ref{fast_fs.table}, the average percent decrease in test error for group $\ell_0$- FAST (top value) is always higher than that for AGIS (bottom value). However, as discussed in \S\ref{agis_vl0.section}, AGIS also has several advantages in terms of speed and ease of use.

For sparsity budget $K = 2$, group $\ell_0$-FAST attains a \textbf{28\%} decrease in test error compared to the best competing algorithm, ControlBurn. Interestingly, the performance of ControlBurn degrades as $K$ increases. This is because the framework selects features independently of the final EBM that is refitted \cite{liu2021controlburn}. While ControlBurn is useful for selecting a few important features, our results show that the framework fails at building sparse PCAMs for larger values of $K$. For $K \in \{4,6,8,10\}$, group $\ell_0$-FAST attains up to a \textbf{30\%} decrease in test error compared to the best competing algorithm, EBM-RS. In all, we observe that group $\ell_0$-FAST consistently outperforms the best competing algorithm across all sparsities.

In addition, feature-sparse FAST substantially outperforms FLAM-GL and FastSparseGAM, by over a \textbf{100\%} decrease in test error for some sparsities. FLAM-GL is affected by over-shrinkage from the group LASSO, which is especially pronounced since the FLAM formulation uses a large number of variables. The $\ell_0$-based penalties and constraints in feature-sparse FAST are shrinkage-free and robust to this effect. FastSparseGAM indirectly selects features by sparsifying piecewise segments in the additive model \citep{liu2022fast}. Feature-sparse FAST, on the other hand, directly accounts for feature sparsity in the optimization framework and outperforms this competing method. We also note that feature-sparse FAST substantially outperforms our competing algorithms that do not fit PCAMs: SAM, which also uses the group LASSO, and the linear LASSO.

 Finally, we observe that many distributions in Figure \ref{violin_plot.fig} have heavy positive tails, notably for the EBM-RS, FLAM-GL, and ControlBurn competing algorithms. These heavy tails typically contain the results from datasets with correlated features; we show correlation matrices and the distribution of errors in the appendix (suppl. F). In \S\ref{semisynthetic.section}, we show through a semi-synthetic experiment that correlated features degrade the performance of  EBM-RS, FLAM-GL, and ControlBurn. Group $\ell_0$-FAST and AGIS, on the other hand, can effectively build sparse PCAMs regardless of feature correlations.

\section{PCAMs and Feature Correlations} 

We conclude by investigating how correlated features impact the interpretability of PCAMs.

\subsection{Correlated Feature Selection}\label{semisynthetic.section}

\begin{figure}[h]
    \centering
    \includegraphics[width = 0.43\textwidth]{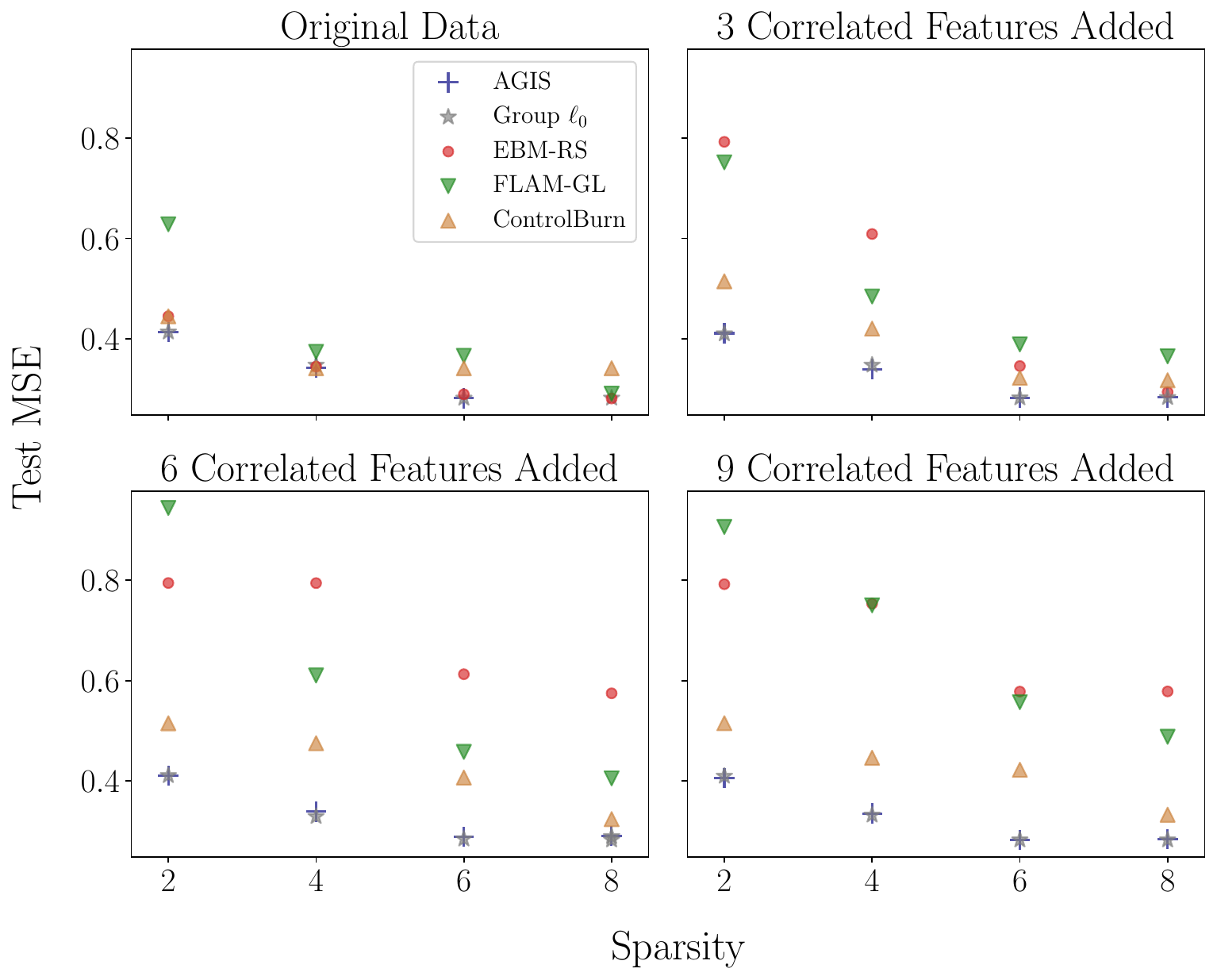}
    \caption{Group $\ell_0$-FAST and AGIS perform well even after adding correlated features.}
    \label{semi_synthetic.fig}
\end{figure}

We present here a semi-synthetic example to investigate how correlated features affect our feature selection experiment (\S\ref{feature_sparse_experiments.section}). We start with the Houses dataset \citep{vanschoren2014openml} and build feature-sparse PCAMs using FAST, EBM-RS, FLAM-GL, and ControlBurn by varying sparsity budget $K$. The top left plot in Figure \ref{semi_synthetic.fig}
shows the test performance of these sparse models.

We then add 3, 6, and 9 correlated features to the data.
As the other plots in Figure \ref{semi_synthetic.fig}  show, the performances of EBM-RS, ControlBurn, and FLAM-GL degrade significantly with added correlations but the performances of group $\ell_0$-FAST and AGIS remain unaffected.

EBM feature importance scores capture the contribution of each feature to the prediction of the model, averaged over all training observations. Given a pair of highly correlated features, the cyclic round-robin algorithm used to fit EBMs will split the contribution of the features evenly between the pair. As such, the average feature importance score/ranking of a group of correlated features will be suppressed, which degrades the performance of EBM-RS. This effect is analogous to the so-called \textit{correlation bias} observed in random forest feature rankings by the ControlBurn paper \cite{liu2021controlburn}.

ControlBurn attempts to address correlation bias by using the weighted LASSO to select features and we indeed observe in Figure \ref{semi_synthetic.fig} that the algorithm is more robust than EBM-RS to added correlations. However, the LASSO penalty used in ControlBurn still imparts shrinkage which biases sparse selection in the presence of multicollinearity \cite{huang2012selective, hastie2020best, mazumder2020discussion}. Additionally, the group LASSO penalty in FLAM-GL is known to suffer from over-shrinkage and performs even worse at selecting sparse subsets of correlated groups \citep{huang2012selective}. The penalties and constraints used to select features in FAST are shrinkage-free and, as a result, our algorithms are unaffected by the added correlated features.





\subsection{Identifying Discontinuities}

One attractive property of PCAMs is their ability to capture discontinuous patterns in the underlying data. Here, we present a case study to demonstrate how correlated features can degrade the ability of EBMs to identify discontinuities.

We use the Houses \cite{vanschoren2014openml} dataset to build PCAMs to predict house prices using demographic features. Unsurprisingly, there is a nearly linear relationship between the median income of a district and the price of homes in that district. We add the following artificial discontinuity to the data: for all districts with a median income above \$40,000 a year, we drop the price of homes by \$20,000.

\begin{figure}[h]
    \centering
    \includegraphics[width = 0.43\textwidth]{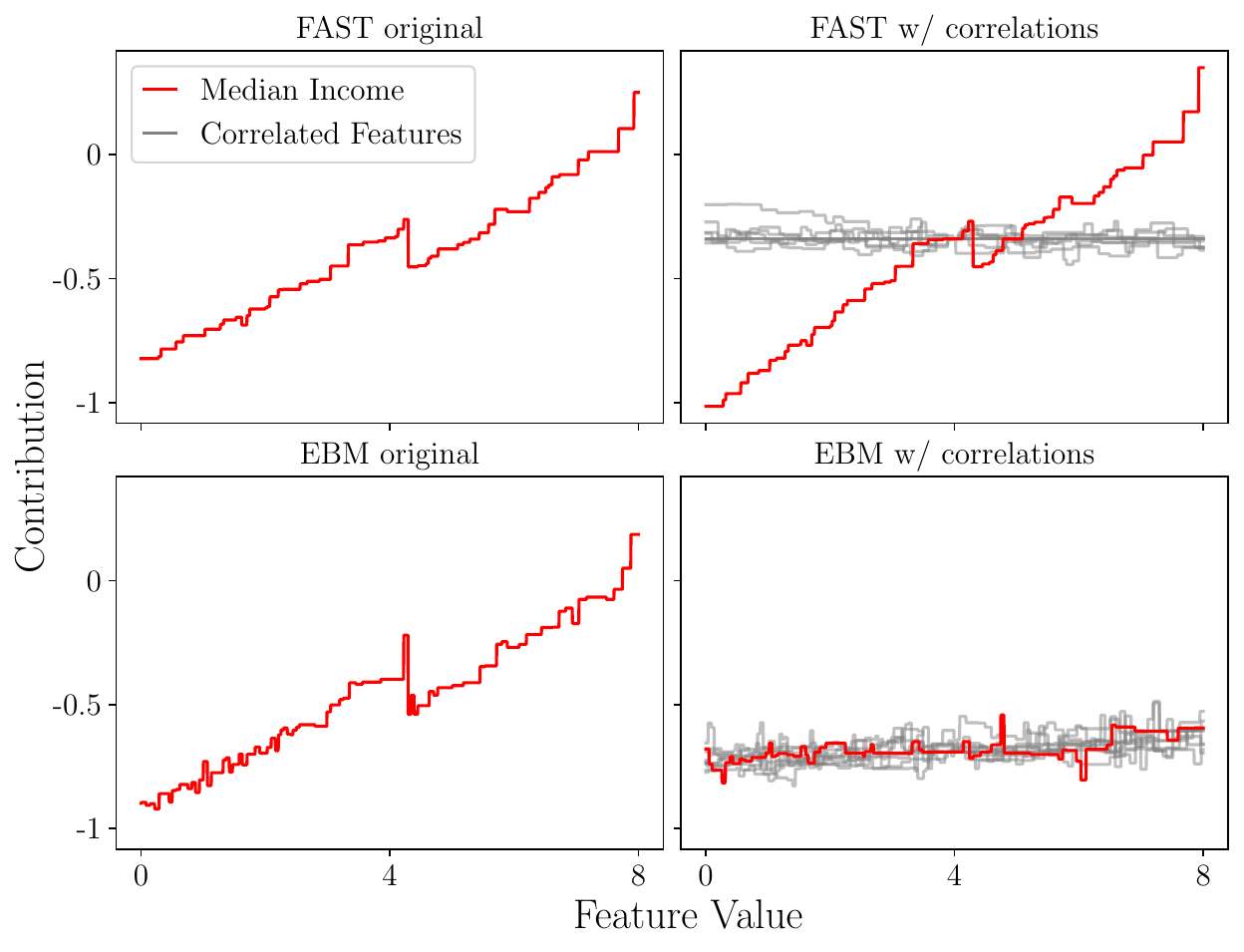}
    \caption{Correlated features can mask discontinuities in PCAM shape functions.}
    \label{interpretability_case.fig}
\end{figure}

We then fit feature-dense PCAMs using FAST and EBMs. Both methods capture the discontinuity in the shape function for median income, as shown in the left two plots in Figure \ref{interpretability_case.fig}. Next, we add 10 synthetic features that are correlated with median income and refit.

The right two plots in Figure \ref{interpretability_case.fig} show the new shape functions. The red lines show median income and the grey lines show the synthetic correlated features. We observe that for FAST (top right), the shape function of median income is preserved and that the discontinuous pattern can still be easily identified. The shape functions of the synthetic correlated features are reduced in magnitude and several are implicitly regularized to zero. The EBM shape functions (bottom right) tell a different story. Due to the cyclic algorithm used to fit EBMs, the contribution of median income is evenly distributed among the noisy correlated features. All the shape functions in this group are compressed and the discontinuity is difficult to detect.


 If a dataset contains a group of correlated features, and only one feature in that group contains an interesting discontinuity, FAST may be more likely to capture this pattern compared to EBMs. Also, consider the case where a dataset contains a sensitive attribute with a discontinuous pattern. A potential adversarial attack would be to add features correlated with this sensitive attribute to mask this discontinuous pattern from EBMs. Due to greedy model fitting, FAST again may be more robust to this attack.

We emphasize that both FAST and EBMs produce transparent PCAMs, however, the interpretations of the models change depending on whether the PCAMs were fit greedily or cyclically. Model transparency does not guarantee trustworthiness and practitioners should still interpret transparent models cautiously.

\vspace{1mm}
\textbf{\noindent Conclusion:} FAST is an optimization-based framework that leverages a novel greedy optimization procedure to fit PCAMs up to 2 orders of magnitude faster than SOTA methods. The framework also introduces two feature selection algorithms that significantly outperform existing methods at building sparse PCAMs. Using FAST, we investigate how correlated features impact the interpretability of PCAMs in terms of selecting important features and interpreting shape functions. These phenomena should be considered when evaluating the trustworthiness of additive models.

\textbf{ACKNOWLEDGMENTS} The authors acknowledge supoort from the ONR (N000142112841, N000142212665, N000142212665, and N000142112841).

\bibliographystyle{plainnat}
\bibliography{ref}
\end{document}